# Adaptive Variable Impedance Control for a Modular Soft Robot Manipulator in Configuration Space

Mahmood Mazare[1], Silvia Tolu[2], Mostafa Taghizadeh[1*]

1- Faculty of Mechanical and Energy Engineering, Shahid Beheshti University, Tehran, Iran.
2- Department of Electrical Engineering, Technical University of Denmark, Kgs. Lyngby, Denmark.
*P.O.B. 1743524155 Tehran, Iran, mo_taghizadeh@sbu.ac.ir

**Abstract**
Compliance is a strong requirement for human-robot interactions. Soft-robots provide an opportunity to cover the lack of compliance in conventional actuation mechanisms, however, the control of them is very challenging given their intrinsic complex motions. Therefore, soft-robots require new approaches to e.g., modeling, control, dynamics, and planning. One of the control strategies that ensures compliance is the impedance control. During the task execution in the presence of coupling force and position constraints, a dynamic behavior increases the flexibility of the impedance control. This imposes some additional constraints on the stability of the control system. To tackle them, we propose a variable impedance control in configuration space for a modular soft robot manipulator (MSRM) in the presence of model uncertainties and external forces. The external loads are estimated in configuration space using a momentum-based approach in order to reduce the calculation complexity, and the adaptive back-stepping sliding mode (ABSM) controller is designed to guard against uncertainties. Stability analysis is performed using Lyapunov theory which guarantees not only the exponential stability of each state under the designed control law, but also the global stability of the closed-loop system. The system performance is benchmarked against other conventional control methods, such as the sliding mode (SM) and inverse dynamics PD controllers. The results show the effectiveness of the proposed variable impedance control in stabilizing the position error and diminishing the impact of the external load compared to SM and PD controllers.

# 1-Introduction

Inspired by living creatures' limbs like the octopus' arm and elephant's trunk, continuum robot manipulators are developed to execute their tasks by continuous deformation of their structure. Their flexibility allows them to undergo large deformations and makes them a suitable choice for dexterous operations in cluttered and confined environments [1–4]. Thus, the contact operation is one of the important aspects to take into account for controlling the interaction with the environment because of its crucial role for a successful task execution.

Theoretically, the interaction task can be successfully performed through pure motion control only when an accurate model of the robot and the environment is available and the parameters are accurately identified. In practice, however, the accurate model of the robot and environment is difficult to obtain. In this regard, controlling the contact force through exclusive motion control of the robot does not yield acceptable results. On the contrary, it is essential to maintain the desired force while tracking the desired trajectories. Un-modeled dynamics, uncertainties, and external disturbances impose a tracking error which in turn adversely affects the interaction forces. Deviation of the contact forces from the desired values can cause the saturation of actuators and increase even more the error, or in extreme cases, it can seriously damage the robot or the environment [5]. To overcome these problems, force control has been applied by numerous researchers.

The first approach to force control is hybrid position/force control in which the task space is decomposed into mutually orthogonal free motions and force spaces and the position and contact forces of the robot are controlled in these spaces. This method mainly focuses on contact forces of the end-effector in task space [6–8]. However, for continuum robots acting in a complex working environment there may be several contact points with surrounding objects. So, this methodology is not an attractive solution. Stiffness control is another method that is exerted to improve the force operation performance of continuum robots, which have a weak structural stiffness [9,10]. This approach evaluates the operation stiffness and forces of the robot by analytically calculating the elastic deformation under certain conditions [9]. However, this approach is configuration-dependent and is not suitable for applications in cluttered and constrained environments, where large configuration changes are not allowed. Impedance control is another method that considers the dynamic relationship between robot interaction forces and motion. It implements a mass-spring compliant system and adjusts the stiffness and damping matrices according to force feedback, so it stably performs the operation task with coupled force-position constraints [11–14]. In other words, compromise between position and force is the main duty of the impedance control strategy. It can be divided into position-based and forced-based methods, each of which is robust against certain type of uncertainties. While force-based impedance control offers a faster response and higher robustness, position-based approach provides more accurate results [15]. Invariant impedance control requires that the stiffness of the environment and its relative location to the robot are known. Thus, for applications with a dynamic contact force tracking in an uncertain environment, it is not an effective solution.

Variable impedance control has been proposed as an effective solution for dynamic force tracking because it compensates uncertainties related to the environment by controlling the dynamic relationship between external forces and robot movements and also can track variable interaction forces. Recently, variable impedance control methodology has gained a numerous attention. In [16], an approach based on variable damping and mass properties of a redundant robot arm was proposed and the stability region of the variable impedance control system was determined. In [17], the stability of variable impedance control was discussed, and a state-independent stability constraint was proposed that relates the stiffness and the time derivative of the stiffness to the damping. In [18], an adaptive variable impedance methodology was employed to control the interaction forces of a commercial industrial robot. The damping parameters of the controller were adaptively adjusted based on the interactive force tracking error. Moreover, the desired speed was tracked in the Cartesian space to achieve interactive force control. In [19], an adaptive variable impedance control was proposed that is capable of tracking dynamic desired forces and compensates for uncertainties related to geometrical and mechanical properties of the environment. The position-based impedance controller actively tracks the dynamic desired force through contact force feedback. In [20], a variable impedance control with time delay estimation was proposed which not only compensates the modeling error and nonlinear dynamics, but also demonstrates good computational efficiency. The impedance parameters were optimized online, using a reinforcement learning algorithm, thereby improving the impedance accuracy and robustness.

Regarding the impedance control of continuum manipulators, few research works were published. In [21], an invariant impedance control was proposed for a single section continuum robot which can deal with unpredicted contacts along the whole manipulator body. In [22], a variable impedance controller was proposed for a cable-driven continuum robot. To simplify the control parameters selection, the authors proposed a simple approach in which a unique scalar parameter was chosen instead of solving a group of nonlinearities.

As stated in [17], the presence of constant elements in the inertia matrix of the impedance system leads to an important and problematic limitation specifically in the task space and in the vicinity of singular configurations. The overriding issue resides in having a prior knowledge of the dynamic model, as well as in sensing the external effort during the interaction with the environment. Hence, the first goal of this study is to provide relative conditions without depending on the system states for time-varying inertia, damping and stiffness matrices of the impedance system. Then, an adaptive robust controller against model uncertainties is designed for the MSRM by using an adaptive back-stepping sliding mode (ABSM). To overcome the challenges of sensing the interaction force, a generalized momentum-based torque estimator is applied.

The organization of this study is as follows: in section 2, the MSRM is introduced and its kinematic and dynamic models are extracted. Section 3 is devoted to the control scheme and its stability analysis. Variable impedance control is presented in section 4. Finally, simulation results, discussion and conclusion are drawn in sections 5, 6, and 7, respectively.

**2-Modular Soft Robot Manipulator**

The MSRM is a flexible and continuum mechanism which is composed of two interconnected and identical modules. Each module includes a combination of three pairs of McKibben-based actuators and three cables which are alternately displaced. Cables and chambers are decoupled, meaning that each one has a dedicated activation line for tension and pressure regulation, respectively. Moreover, cables activity is decoupled along the two modules thanks to the use of Bowden cables. Fig.1 shows an overall overview of the MSRM.

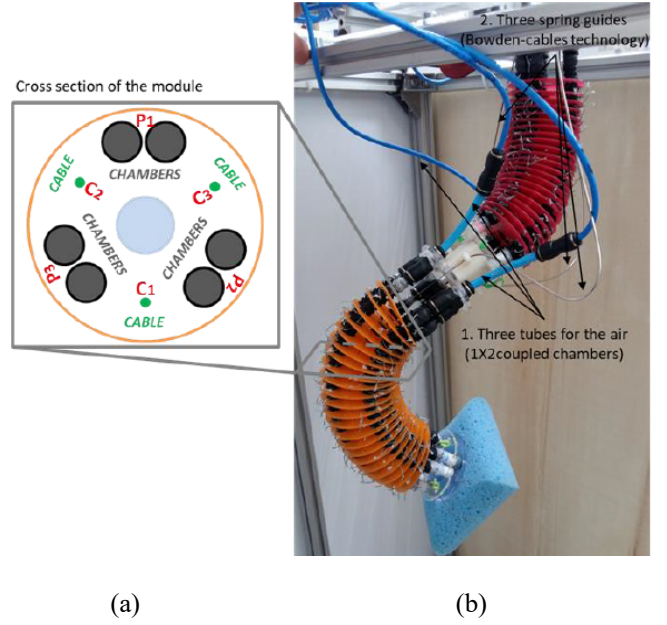

(a)          (b)

**Fig.1. (a) Overall design of a single module with a section view showing the arrangement of the actuators. P1, P2, P3 are for each pair of pneumatic chambers while C1, C2 and C3 mean actuation cables;(b) physical prototype of the soft manipulator.**

**2.1- Kinematic model**

Kinematic model represents the coordinates of the robot end-effector in task space by using the actuator states. Different coordinate frames including the actuator space, configuration space, and task space are considered in order to extract the kinematic model. Different states of these frames are as:

Actuator states:
$$q = \begin{bmatrix} l_{i,1} & l_{i,2} & l_{i,3} \end{bmatrix}^T \quad i=1,2$$

Configuration states:
$$\Omega = \begin{bmatrix} \kappa_i & \phi_i \end{bmatrix}^T \quad i=1,2 \quad (1)$$

Task states:
$$\chi = \begin{bmatrix} x & y & z & \omega_x & \omega_y & \omega_z \end{bmatrix}^T$$

where $l_i$ denotes the effective length of the cables in each segment. Curvature $\kappa_i$ and torsion angle $\phi_i$ can be defined as a function of actuator states, and the dihedral angle is defined as $\theta_i = \kappa_i l_i$. The relation between each pair of the states is depicted in Fig.2.

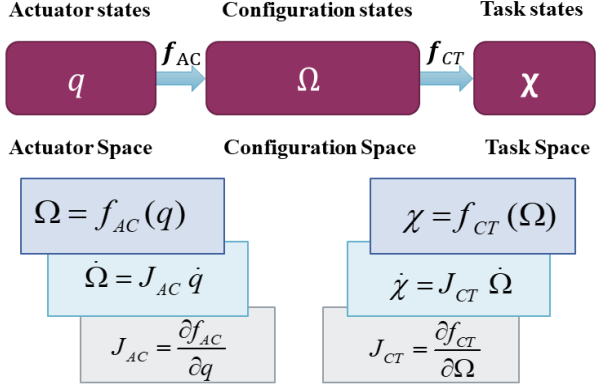

**Fig. 2 Kinematic spaces**

Task space coordinates comprise both rotational and translational coordinates and can be determined through a homogeneous transformation matrix. The transformation includes a rotation matrix and a translation vector. The relative posture between each pair of the coordinate frames as well as derivation of any task coordinates such as position can be determined by the homogeneous transformation matrix. Next, kinematic transformations will be described.

- **Kinematic transformations**

Based-on the different coordinate frames, determination of mapping between each pair of these frames can be performed (Fig.2). From the figure, the first mapping ($f_1$) transforms the actuator states into configuration states and the second mapping ($f_2$) determines the task states using configuration states. In other words, configuration states can be obtained through $f_1$ which is a function of actuator states and based on constant curvature approach, the mapping can be determined. The second mapping between the base and tip of each section can be determined using the homogeneous transformation matrix which is defined as:

$$^{j}H_i = \begin{bmatrix} ^{j}R_i & ^{j}O_i \\ 0_{1\times 3} & 1 \end{bmatrix}$$

$$^{ib}H_{it} = \begin{bmatrix} c_\phi^2 a + 1 & s_\phi c_\phi a & s_\phi c_{\kappa l} & \dfrac{c_\phi a}{\kappa} \\ s_\phi c_\phi a & c_{\kappa l} - c_\phi^2 a & s_\phi s_{\kappa l} & \dfrac{-s_\phi a}{\kappa} \\ -c_\phi s_{\kappa l} & -s_\phi s_{\kappa l} & c_{\kappa l} & \dfrac{s_{\kappa l}}{\kappa} \\ 0 & 0 & 0 & 1 \end{bmatrix} \quad (2)$$

where $a = (c_{\kappa l} - 1)$, $c_\bullet = \cos(\bullet_i)$, $s_\bullet = \sin(\bullet_i)$, $t$ and $b$ denote tip or head, and base of a single section. Also, $i$ and $j$ represent coordinate frames. By using the different coordinate frames, kinematic model can be determined. Assuming a constant curvature, the specific robot configuration can be expressed as a function of the actuator states:

$$\kappa_i = 2\dfrac{\sqrt{l_{i,1}^2 + l_{i,2}^2 + l_{i,3}^2 - l_{i,1}l_{i,2} - l_{i,2}l_{i,3} - l_{i,3}l_{i,1}}}{r(l_{i,1} + l_{i,2} + l_{i,3})}$$

$$\phi_i = \tan^{-1}\left(\dfrac{2l_{i,1} - l_{i,2} - l_{i,3}}{\sqrt{3}(l_{i,2} - l_{i,3})}\right) \quad (3)$$

$$l_i = \dfrac{(l_{i,1} + l_{i,2} + l_{i,3})}{3}$$

where $r$ denotes the distance of the disks' center to the section's backbone. The position vector ($r$) in tip (index $t$) of a section $i$ can be transformed to its base (index $b$) coordinate frame, and the world reference frame ($O_W$) as follows:

$$^{b_i}r_i = {}^{b_i}H_{t_i} \bullet {}^{t_i}r_i$$
$$^{O_w}r_i = {}^{O_w}H_{t_i} \bullet {}^{t_i}r_i \quad (4)$$

Moreover, the Jacobian between each pair of these frames can be calculated as $J_{AC} = \partial f_{AC}/\partial q$, $J_{CT} = \partial f_{CT}/\partial \Omega$ and $J = J_{AC}J_{CT}$. The schematic of the MSRM is shown in Fig.3.

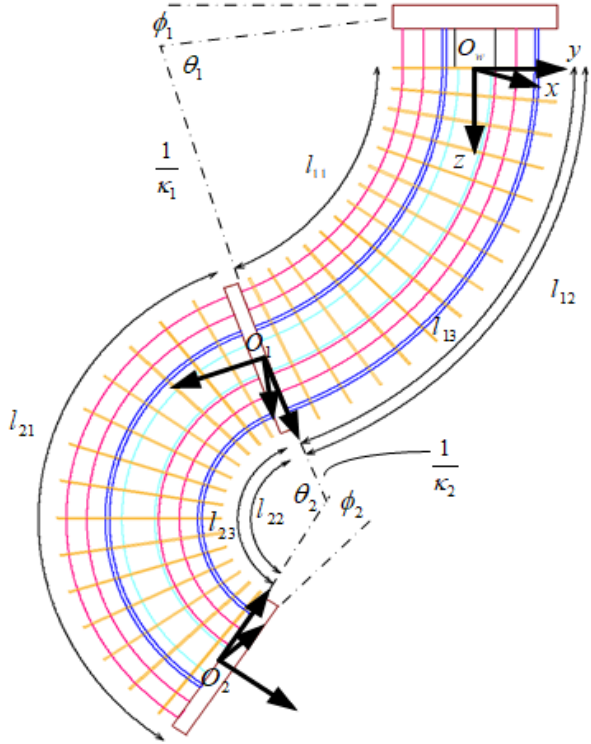

**Fig. 3 CAD model of the MSRM**

## 2.2-The Dynamic model

In this section, dynamic model of the MSRM will be extracted based on pseudo-rigid-body model (Fig.4). For modeling a dynamic system using Euler-Lagrange formulation, the first step is to determine the Lagrangian, which is defined by

$$L = T - U \quad (5)$$

where $T$ and $U$ represent kinetic and potential energy respectively. Here, the only conservative force stored in the form of potential energy is gravity.

$$\begin{aligned} U &= U_1 + U_2 \\ U_1 &= \frac{1}{2} k_\phi \phi^2 + \frac{1}{2} k_\psi \psi^2 \\ U_2 &= -m\, g\, z \end{aligned} \quad (6)$$

The kinetic energy of the MSRM is obtained as:

$$\begin{aligned} T &= T_1 + T_2 \\ T_1 &= \frac{1}{2} J_\phi \dot{\phi}^2 + \frac{1}{2} J_\psi \dot{\psi}^2 \\ T_2 &= \frac{1}{2} m\dot{x}^2 + \frac{1}{2} m\dot{y}^2 + \frac{1}{2} m\dot{z}^2 \end{aligned} \quad (7)$$

By substituting kinetic and potential energy into Eq.5 and using Euler-Lagrange formulation, the general dynamic model of the soft robot can be expressed in the following standard form:

$$M(\Omega)\ddot{\Omega} + C(\Omega,\dot{\Omega})\dot{\Omega} + N(\Omega) = \tau_e + \tau_c \quad (8)$$

where $M(\Omega)$, $C(\Omega,\dot{\Omega})$ and $N(\Omega)$ represent the inertia, Coriolis, and potential terms respectively. Moreover, $\tau_e$ is the external load and $\tau_c$ denotes the actuation torque.

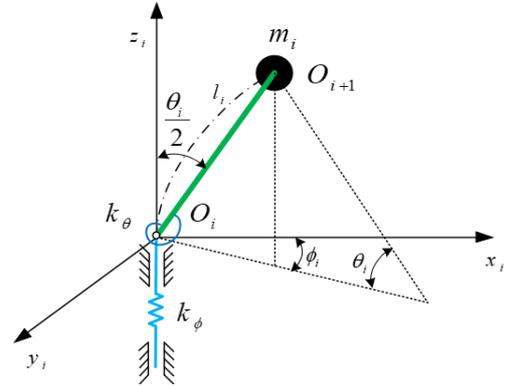

**Fig. 4 Pseudo-rigid-body model of an elastic segment**

## 3- Control strategy

In this section, an adaptive back-stepping controller will be designed for the understudied soft robot. For the sake of comparison, two nonlinear controllers, sliding mode (SM) control law and proportional derivative (PD) - based feedback linearization law, are simulated for the MSRM.

### 3.1- Adaptive Back-stepping Sliding Mode Control

In this section, an ABSM control law will be designed based on the dynamic model derived in the previous section. A state-space representation of the MSRM model Eq.8 is as following:

$$\begin{cases} \dot{X}_1 = X_2 \\ \dot{X}_2 = \bar{M}^{-1}\left[\tau_c - \bar{C}X_2 - \bar{N} + \tau_e\right] \end{cases} \quad (9)$$

**Assumption 1**: dynamic model parameters can be described as:

$$\begin{cases} M = \bar{M} + \Delta M \\ C = \bar{C} + \Delta C \\ N = \bar{N} + \Delta N \end{cases} \quad (10)$$

where $\bar{M}$, $\bar{C}$ and $\bar{N}$ represent the nominal part, and $\Delta M$, $\Delta C$ and $\Delta N$ denote the model uncertainties. Then, the dynamic model can be rewritten as:

$$\begin{cases} \dot{X}_1 = X_2 \\ \dot{X}_2 = M^{-1}\left[\tau_c - CX_2 - N + D + \tau_e\right] \end{cases} \quad (11)$$

in which $D = \Delta M \dot{X}_2 + \Delta C X_2 + \Delta N$ expresses the lumped uncertainty term.

**Assumption 2:** it is assumed that modelling errors and external disturbances in the robot manipulator system change slowly, that is $\dot{D} = 0$.

To design an adaptive back-stepping sliding mode control law, the main steps are introduced as follows.

- **Step 1:**

According to the tracking objective for the robotic systems, the trajectory tracking error, virtual control variables and the velocity error are defined as:

$$\begin{aligned} e_p &= \Omega - \Omega_d \\ \sigma &= \varepsilon e_p \\ e_v &= \dot{e}_p + \sigma \end{aligned} \quad (12)$$

where $\varepsilon$ is a positive definite diagonal coefficient matrix. The following Lyapunov function is selected as a function of position error:

$$V_1 = \tfrac{1}{2} e_p^T e_p \quad (13)$$

Taking the time derivative of the Lyapunov function leads to:

$$\begin{aligned} \dot{V}_1 &= e_p^T \dot{e}_p \\ &= e_p^T (e_v - \varepsilon e_p) \\ &= e_p^T e_v - e_p^T \varepsilon e_p \end{aligned} \quad (14)$$

If $e_v = 0$, then $\dot{V}_1 \leq 0$. To show the negative property for the first Lyapunov function, we will go to the next step.

- **Step 2:**

A simple sliding surface is chosen as $s = \lambda e_p + e_v$ in which $\lambda$ is a positive constant. Next, the corresponding Lyapunov function is selected as:

$$V_2 = V_1 + \tfrac{1}{2} s^T s \quad (15)$$

Then, its time derivative is obtained as:

$$\begin{aligned} \dot{V}_2 &= e_p^T \dot{e}_v - e_p^T \varepsilon e_p + s^T \lambda (e_v - \varepsilon e_p) \\ &+ s^T \left[ M^{-1}(\tau - CX_2 - N - D) - \ddot{\Omega}_d + \varepsilon \dot{e}_p \right] \end{aligned} \quad (16)$$

To satisfy $\dot{V}_2 \leq 0$, the following control law is defined:

$$\begin{aligned} \tau_c &= M\left[-\lambda(e_v - \varepsilon e_p) + \ddot{\Omega}_d - \varepsilon \dot{e}_p - s\right] \\ &+ CX_2 + N + D - \tau_e - \gamma M \tanh(s/\delta) \end{aligned} \quad (17)$$

- **Step 3:**

In a real-time control system, it is difficult to predict the general uncertainty, $D$. To avoid adopting the upper bounds of $D$, an adaptive algorithm is applied based on back-stepping sliding mode control. According to Assumption 2 and by defining $\tilde{D} = D - \hat{D}$, we have $\dot{\tilde{D}} = \dot{D} - \dot{\hat{D}} = -\dot{\hat{D}}$. The third quadratic Lyapunov function is defined as:

$$V_3 = V_2 + \tfrac{1}{2} \eta^{-1} \tilde{D}^T \tilde{D} \quad (18)$$

where $\eta$ is a positive constant. The time derivative of $V_3$ can be expressed as:

$$\begin{aligned} \dot{V}_3 &= e_p^T \dot{e}_v - e_p^T \varepsilon e_p - \kappa^{-1} \tilde{D}^T \left(\dot{\hat{D}} + \eta M^{-T} s\right) \\ &+ s^T \left[(\varepsilon \dot{e}_p - \ddot{\Omega}_d) + \lambda(e_v - \varepsilon e_p)\right] \\ &+ s^T \left[M^{-1}(\tau_c - CX_2 - N - D + \tau_e)\right] \end{aligned} \quad (19)$$

By choosing the adaptation law as $\dot{\hat{D}} = -\eta M^{-T} s$ the ABSM control law is derived as:

$$\tau_c = M\left[-\lambda(e_v - \varepsilon e_p) + \ddot{\Omega}_d - \varepsilon \dot{e}_p - \nu s\right]$$
$$+C X_2 + N + \hat{D} - \tau_e - \gamma M \nu \tanh(s/\delta) \quad (20)$$
$$\dot{\hat{D}} = -\eta M^{-T} s$$

where $\gamma$ and $\nu$ are positive constants.

**Theorem 1:** The MSRM system is asymptotically stable with the proposed adaptive robust control law (Eq.20).

**Proof:** Considering the third Lyapunov function (Eq.18) and its time derivative (Eq.19) and substituting the control law (Eq.20) into Eq.19 yields:

$$\begin{aligned}\dot{V}_3 &= e_p^T e_v - e_p^T \varepsilon e_p - s^T \nu s - \gamma s^T \nu \tanh(s/\tau) \\ &= e_p^T e_v - e_p^T \varepsilon e_p - \nu\|s\|^2 - \gamma \nu\|s\| \\ &\leq e_p^T e_v - e_p^T \varepsilon e_p - s^T \gamma \nu s\end{aligned} \quad (21)$$

Next, for the sake of simplification, Eq.21 will be rewritten in matrix form. By defining $E = \begin{bmatrix} e_p & e_v \end{bmatrix}$ and $\Phi$ as a positive definite matrix:

$$\Phi = \begin{bmatrix} \varepsilon + \lambda^T \nu \lambda & \nu\lambda - 0.5 \\ \nu\lambda - 0.5 & \nu \end{bmatrix}$$
$$E^T \Phi E = e_p^T \varepsilon e_p - e_p^T e_v + s^T \nu s \quad (22)$$
$$\|\Phi\| = \nu(\varepsilon + \lambda) - 0.25$$

by choosing proper parameters for $\nu$, $\varepsilon$ and $\lambda$ to satisfy $\|\Phi\| \geq 0$, the transformation matrix will be a positive definite matrix. Then, Eq. 21 can be rewritten as

$$\dot{V}_3 \leq -E^T \Phi E \quad (23)$$

which means that $\dot{V}_3 \leq 0$. Therefore, the control loop will be stable and the tracking error will asymptotically converge to zero with respect to the third quadratic Lyapunov function. This completes the proof.

### 3.2- Sliding mode control law

In this part, a robust sliding mode control law is designed. By defining a sliding surface as $s = \lambda e_p + e_v$ and by taking time derivative of the surface ($\dot{s} = 0$), then the sliding mode control law can be derived as [24]:

$$\tau_{c_{SM}} = M\left[-\lambda(e_v - \varepsilon e_p) + \ddot{\Omega}_d - \varepsilon \dot{e}_p - s\right]$$
$$+C X_2 + N - \tau_e - \gamma M \tanh(s/\delta) \quad (24)$$

### 3.3- Proportional derivate based feedback linearization control law

Feedback linearization technique using inverse dynamics PD is implemented in tracking control of the proposed mechanism. The following control law is applied in the controller.

$$u = \ddot{\Omega}_d - K_D \dot{e}_p - K_P e_p \quad (25)$$

which $K_D$ and $K_P$ are control parameters. Substituting $\ddot{\Omega} = u$ in the inverse dynamics equation, the computed actuator torques can be determined. This control action leads to error dynamics in the form of:

$$\ddot{e}_p + K_D \dot{e}_p + K_P e_p = 0 \quad (26)$$

To satisfy the exponential convergence of the general system, $K_D$ and $K_P$ should be positive definite [24].

## 4- Variable Impedance Control

Based on the dynamic model derived in the previous section, a variable impedance control strategy will be presented in the configuration space.

### 4.1- Impedance Control

The main objective of the impedance control in configuration space is to develop a dynamic relation between the external force and configuration error, as [15]:

$$M_d\left(\ddot{\Omega} - \ddot{\Omega}_d\right) + C_d\left(\dot{\Omega} - \dot{\Omega}_d\right) + K_d\left(\Omega - \Omega_d\right) = \tau_e \quad (27)$$

in which $M_d$, $C_d$, and $K_d$ are the desired inertia, damping and stiffness respectively. These matrices are adjusted by the user in order to determine the behavior of the soft robot against external forces. It should be noted that asymptotic stability of the system will be guaranteed if theses parameters are constant.

**Theorem 2:** We consider $M_d$, $C_d$, and $K_d$ as symmetric and positive definite matrices and also continuously differentiable varying profiles. It can be proved that the system Eq.27 when $\tau_e = 0$ is uniformly globally asymptotically stable (UGAS) if there exist a positive constant $\alpha$ so that:

$$\begin{cases} \dot{M}_d + \alpha M_d - C_d \leq 0 \\ (\alpha^2 + 2\alpha)\dot{M}_d - \alpha \ddot{M}_d + \alpha \dot{C}_d + \dot{K}_d - 2\alpha K_d \leq 0 \end{cases} \quad (28)$$

**Proof:** In order to perform the stability analysis, we use a following Lyapunov function:

$$V = \frac{1}{2}\left(\Upsilon^T M_d \Upsilon + \Xi^T \mu \Xi\right) \quad (29)$$

where $\mu$ is a symmetric and positive-definite matrix and $\Upsilon = \dot{\Xi} + \alpha \Xi$, $\Xi = \Omega - \Omega_d$. By taking time derivative of Eq. 29 as:

$$\dot{V} = \Upsilon^T M_d \left(\ddot{\Xi} + \alpha \dot{\Xi}\right) + \frac{1}{2}\Upsilon^T \dot{M}_d \Upsilon \\ + \Xi^T \mu \dot{\Xi} + \frac{1}{2}\Xi^T \dot{\mu} \Xi \quad (30)$$

when $\tau_e = 0$,

$$\dot{V} = \dot{\Xi}^T \left(\dot{M}_d + \alpha M_d - C_d\right)\dot{\Xi} \\ + \dot{\Xi}^T \left(\alpha^2 M_d + \alpha \dot{M}_d - \alpha C_d - K_d + \mu\right)\Xi \quad (31) \\ + \Xi^T \left(\alpha^2 \dot{M}_d + \alpha^2 M_d + \frac{1}{2}\dot{\mu} - \alpha K_d\right)\Xi$$

By defining $\mu = -\alpha^2 M_d - \alpha \dot{M}_d + K_d + \alpha C_d$ and $\dot{\mu} = -\alpha^2 \dot{M}_d - \alpha \ddot{M}_d + \alpha \dot{C}_d + \dot{K}_d$, then:

$$\dot{V} = \dot{\Xi}^T \left(\dot{M}_d + \alpha M_d - C_d\right)\dot{\Xi} \\ + \Xi^T \begin{bmatrix} \left(\frac{\alpha^2}{2} + \alpha\right)\dot{M}_d - \frac{\alpha}{2}\ddot{M}_d + \frac{\alpha}{2}\dot{C}_d \\ + \frac{1}{2}\dot{K}_d - \alpha K_d \end{bmatrix} \Xi \quad (32)$$

in order to stabilize the system, the following inequality must be hold:

$$\begin{cases} \dot{M}_d + \alpha M_d - C_d \leq 0 \\ (\alpha^2 + 2\alpha)\dot{M}_d - \alpha \ddot{M}_d + \alpha \dot{C}_d + \dot{K}_d - 2\alpha K_d \leq 0 \end{cases} \quad (33)$$

which completes the proof.

The conditions defined in Eq.33 impose constraints on the impedance matrices which can be determined using an optimization procedure. In order to satisfy the necessary conditions for Theorem 1 and also least conservative constraints, $\alpha$ should be determined using:

$$\alpha = \min\left(\frac{\lambda_{\min}(C_d) + \lambda_{\max}(\dot{M}_d)}{\lambda_{\max}(M_d)}\right) \quad (34)$$

To demonstrate the systematic procedure, let us define a symmetric and negative definite matrix as $B = \dot{M}_d + \alpha M_d - C_d$ whose terms comprise symmetric matrices. For negativity of this matrix, its maximum eigenvalue $\lambda_{\max}$ must be negative. Based on the triangle inequality for supremum norm we have:

$$\lambda_{\max}(B) = \sup_{\|v\|=1} v^T B v \leq \sup_{\|v\|=1} v^T \dot{M}_d v \\ + \sup_{\|v\|=1} v^T (\alpha M_d) v + \sup_{\|v\|=1} v^T (-C_d) v \quad (35) \\ \lambda_{\max}(B) \leq \lambda_{\max}(\dot{M}_d) + \lambda_{\max}(\alpha M_d) - \lambda_{\min}(C_d)$$

and for satisfying the negative definiteness of B, $\lambda_{\max}(\dot{M}_d) \leq \lambda_{\min}(C_d) - \alpha \lambda_{\max}(M_d)$. We move to the second constraint that imposes a limitation on the variation of stiffness matrix. Let us consider $Q = (\alpha^2 + 2\alpha)\dot{M}_d - \alpha \ddot{M}_d + \alpha \dot{C}_d + \dot{K}_d - 2\alpha K_d$ and similar to the first constraint using the triangle inequality, then

$$\lambda_{\max}(\dot{K}_d) \leq 2\alpha \lambda_{\min}(K_d) - \alpha \lambda_{\max}(\dot{C}_d) - (\alpha^2 + 2\alpha) \\ \lambda_{\min}(C_d) + \alpha^2(\alpha + 2)\lambda_{\max}(\dot{M}_d) + \alpha \lambda_{\min}(\ddot{M}_d) \quad (36)$$

Hence, the validity of the impedance profiles can be guaranteed through these conditions, which completes the proof of the stability of varying impedance control without dependency on state measurements.

### 4.2- Estimation of external load torques

Load estimation can facilitate sensing the interaction forces compared to traditional approach in which a multi-dimensional torque sensor should be attached to the end-effector of the robot. Thus, in this paper, a torque estimator is employed to create the closed-loop controller for the MSRM. To realize variable impedance control in configuration space, this approach may be the unique feasible method at this moment. The external load is estimated based on the generalized momentum of a soft manipulator system [25-26]. Using the pseudo-rigid-body dynamic modeling approach, the generalized momentum of the MSRM and its time derivative can be expressed as:

$$P = M(\Omega)\dot{\Omega} \qquad (37)$$
$$\dot{P} = M(\Omega)\ddot{\Omega} + \dot{M}(\Omega)\dot{\Omega}$$

Since $\dot{M}(\Omega) - 2C(\Omega,\dot{\Omega})$ is a skew symmetry matrix for the derived dynamic model, then:

$$\dot{M}(\Omega) = C(\Omega,\dot{\Omega}) + C^T(\Omega,\dot{\Omega}) \qquad (38)$$

By substituting Eq.38 and the dynamic model of the MSRM into Eq.37, time derivative of the generalized momentum is rewritten as:

$$\begin{aligned}\dot{P} &= \tau_e + \tau_c - C(\Omega,\dot{\Omega})\dot{\Omega} - N(\Omega) \\ &\quad + C(\Omega,\dot{\Omega})\dot{\Omega} + C^T(\Omega,\dot{\Omega})\dot{\Omega} \\ &= \tau_e + \tau_c - N(\Omega) + C^T(\Omega,\dot{\Omega})\dot{\Omega}\end{aligned} \qquad (39)$$

By using Eq.38, the external load can be estimated through the following theorem.

**Theorem 3:** the residual vector can be defined in the following form [26]:

$$r(t) = K_I\left[P(t) - \int(\tau_c - N(\Omega) + r(s))ds - P(0)\right] \qquad (40)$$

where $r(0) = 0$ and $K_I > 0$. During slow motion, the dynamic system (Eq.39) can be approximated as:

$$\dot{r}(t) \approx -K_I\left[r(t) - \tau_e\right] \qquad (41)$$

when $K_I \to +\infty$, then $\dot{r}(t) \approx \tau_e$.

**Proof:** using Eq.39, time derivative of the residual vector can be expressed as:

$$\begin{aligned}\dot{r}(t) &= K_I\begin{bmatrix}\tau_e + \tau_c - N(\Omega) + C^T(\Omega,\dot{\Omega})\dot{\Omega} \\ -\tau_c + N(\Omega) - r(t)\end{bmatrix} \\ &= -K_I\left[r - \tau_e - C^T(\Omega,\dot{\Omega})\dot{\Omega}\right]\end{aligned} \qquad (42)$$

It can be shown that centrifugal force is smaller than the inertia force [23]. So, the third term, $K_I C^T(\Omega,\dot{\Omega})\dot{\Omega}$ can be neglected. The solution of the differential equation is as:

$$r(t) \approx r(0)e^{-K_I t} + e^{-K_I t}\int_0^t K_I \tau_e e^{K_I s}ds \qquad (43)$$

According to the initial state $r(0) = 0$, then Eq.43 can be rewritten as:

$$\begin{aligned}r(t) &\approx \int_0^t K_I \tau_e e^{-K_I(t-s)}ds \\ &= \tau_e \int_0^t d\left(e^{-K_I(t-s)}\right)e^{-K_I(t-s)} \\ &= \tau_e\left(1 - e^{-K_I t}\right)\end{aligned} \qquad (44)$$

Hence, the external load can be approximately estimated using Eq.39 when $K_I \to +\infty$. This completes the proof.

### 4.3- Impedance Control in Configuration Space

The robot must work in task space in practical applications. In this section, impedance control using the transformation between task space and configuration space will be developed. The variable impedance control in task space can be introduced as:

$$\hat{M}_d(\ddot{\chi} - \ddot{\chi}_d) + \hat{C}_d(\dot{\chi} - \dot{\chi}_d) + \hat{K}_d(\chi - \chi_d) = F_{ext} \qquad (45)$$

in which $\chi$ represents the position vector of the end-effector in task space and $F_{ext} \in R^{6\times 1}$ indicates the external forces acting on the robot. $\hat{M}_d \in R^{6\times 6}$ is a constant diagonal matrix, while $\hat{C}_d \in R^{6\times 6}$ and $\hat{K}_d \in R^{6\times 6}$ are the time-varying impedance matrices.

Using different transformations between joint space, configuration space, and task space, $\dot{\chi} = J_1 \dot{\Omega}, \ddot{\chi} = J_1 \ddot{\Omega} + \dot{J}_1 \dot{\Omega}$ and also $\dot{\chi} = \ddot{\chi} = 0$, Eq.45 can be rewritten as:

$$\hat{M}_d J_1 \ddot{\Omega} + \left(\hat{C}_d J_1 + \hat{M}_d \dot{J}_1\right)\dot{\Omega} + \hat{K}_d \left(\chi - \chi_d\right) = F_{ext} \quad (46)$$

Note that $\chi = \chi_d$, then $(\chi - \chi_d) = J_1(\Omega - \Omega_d)$ and considering $\ddot{\Omega}_d = \dot{\Omega}_d = 0$ in impedance control system, multiplying both side of Eq.46 by $J_1^T$ it yields:

$$\begin{aligned} M_c \ddot{\Omega} + C_c \dot{\Omega} + K_c (\Omega - \Omega_d) &= \tau_{ext} \\ M_c = J_1^T \hat{M}_d J_1, \ C_c &= J_1^T \hat{M}_d \dot{J}_1 + J_1^T \hat{C}_d J_1 \\ K_c = J_1^T \hat{K}_d J_1, \tau_{ext} &= J_1^T F_{ext} \end{aligned} \quad (47)$$

Therefore, by adjusting the impedance matrices of Eq.45, the local stability of Eq.47 can be guaranteed.

**5- Results**

In this section, the closed-loop system of the robot comprising adaptive back-stepping sliding mode control combined with variable impedance control is simulated, and the obtained results are presented. For the sake of comparison and to show the robustness of the controller against uncertainties and disturbances, the simulations are also conducted for the systems employing the inverse dynamics PD control and conventional SM control. Moreover, to reveal the effectiveness of the variable impedance control, a simulation is also performed by applying invariable impedance control. The control parameters are presented in table 1, and the transformed initial condition from task space to configuration space is $\begin{bmatrix} 0.02 & -0.01 & 0.01 & -0.03 \end{bmatrix}$. Considering the impedance model as Eq.45, it is assumed that the desired inertia, damping and stiffness for both invariable and variable impedance control strategies are defined as presented in table.2.

Table 1. controller parameters

| Parameter | Value |
|---|---|
| $K_P$ | 125 |
| $K_D$ | 5 |
| $\varepsilon$ | 2.9 |
| $\nu$ | 1.9 |
| $\gamma$ | 1.1 |
| $\lambda$ | 10.5 |
| $\delta$ | 0.05 |
| $\eta$ | 1000 |

Table.2. Numerical values for impedance matrices

| Symbols | Variable Imp. | Invariable Imp. |
|---|---|---|
| $\hat{M}_d$ | $\left[15 + 10\sin\left(\frac{\pi}{5}t\right)\right]I_{6\times 6}$ | $15 I_{6\times 6}$ |
| $\hat{C}_d$ | $\dot{\hat{M}}_d + \alpha \hat{M}_d$ | $20 I_{6\times 6}$ |
| $\hat{K}_d$ | $\left[30 + 20\sin\left(\frac{\pi}{2}t\right)\right]I_{6\times 6}$ | $30 I_{6\times 6}$ |

Additionally, it is assumed that three distinct interaction forces, as illustrated in Fig. 5, are applied to the end segment of the robot.

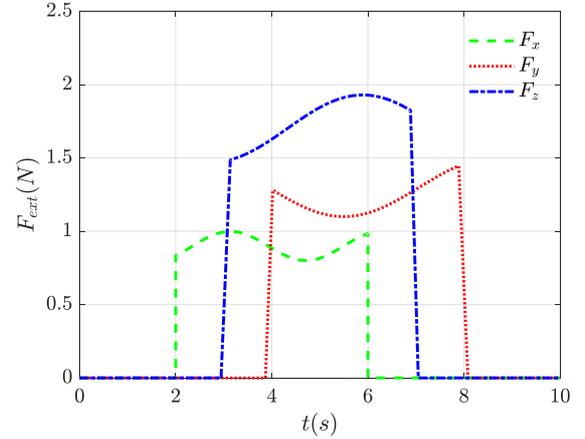

**Fig. 5 External forces applied to the robot end point as pulse signals**

Numerical values of physical parameters of the robot are presented in Table 3.

Table 3. Structural parameters of the MSRM

| Parameters | Symbols | Value | Unit |
|---|---|---|---|
| Radius of the disks | $R_i$ | 0.03 | m |
| Arc length of single segment | $l_i$ | 0.15 | m |
| Mass | $m_i$ | 0.25 | Kg |
| Bending stiffness | $k_\psi$ | 0.5 | Nm |
| Torsional stiffness | $k_\phi$ | 1.0 | Nm |
| Bending inertia | $J_\psi$ | $4.5\times 10^{-3}$ | Kgm$^2$ |
| Torsional inertia | $J_\phi$ | $9\times 10^{-4}$ | Kgm$^2$ |

Measurement of the external loads is a challenging task in reality due to the special structure of soft robots. To overcome this issue, a momentum-based load estimator is designed to estimate these loads. The external forces are mapped to the configuration space as depicted in Fig 6. As can be seen from the figure, the torques corresponding to curvatures are much more considerable, accounting for around 0.12 N.m. Thus, for the sake of brevity, exclusively the estimated values for these components are presented. Obviously, as the constant coefficient $K_I$ increases, the accuracy of the estimation law (Eq.44) is enhanced, showing more compliance with the external load in configuration space.

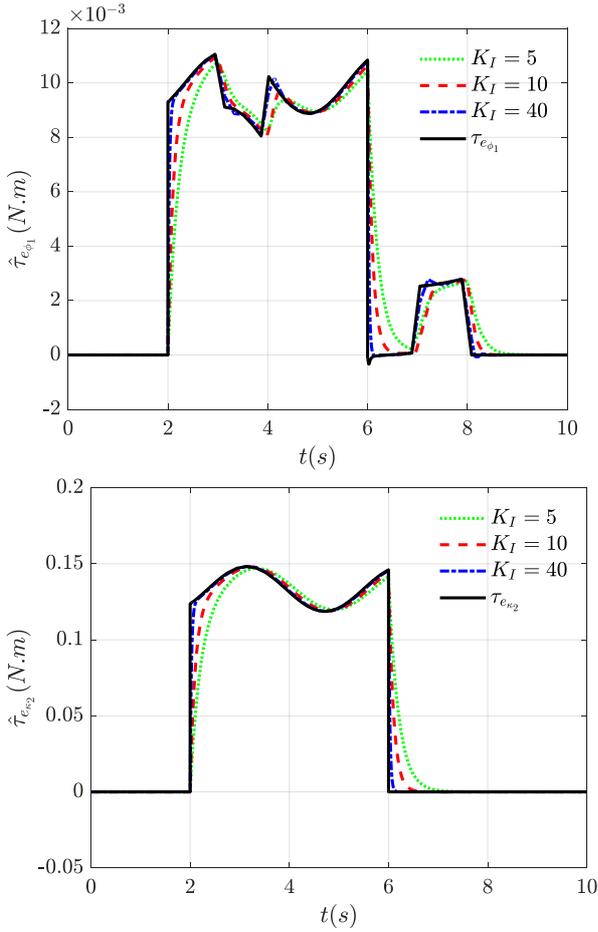

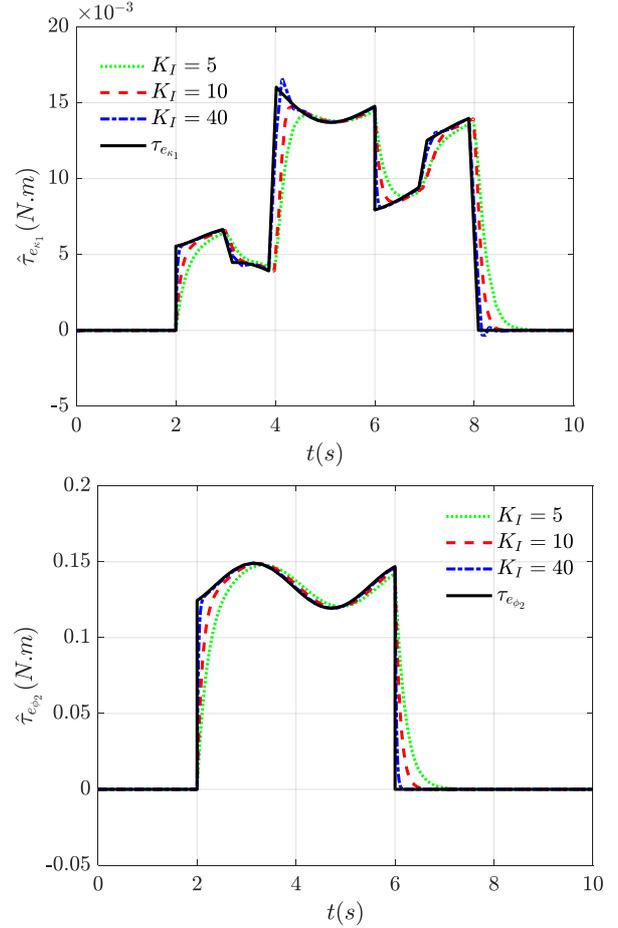

Fig. 6 Estimated external load components ($\tau_{e_{\kappa_1}}$, $\tau_{e_{\phi_1}}$, $\tau_{e_{\kappa_2}}$, $\tau_{e_{\phi_2}}$) in configuration space

Fig. 7 shows the position error of the end-effector under the influence of interaction forces for the proposed variable impedance control strategy, and compares the results with those obtained from conventional SM and PD control strategies. As discussed earlier, impedance control tries to make a compromise between position and force errors. This is the reason behind the increased position errors between *2 sec*. and *8 sec.*, the time span in which the external forces are applied to the robot. In spite of the fact that all controllers have maintained the stability of the closed loop system, the proposed controller has demonstrated far better robustness, showing nearly zero position errors even under the effect of external loads.

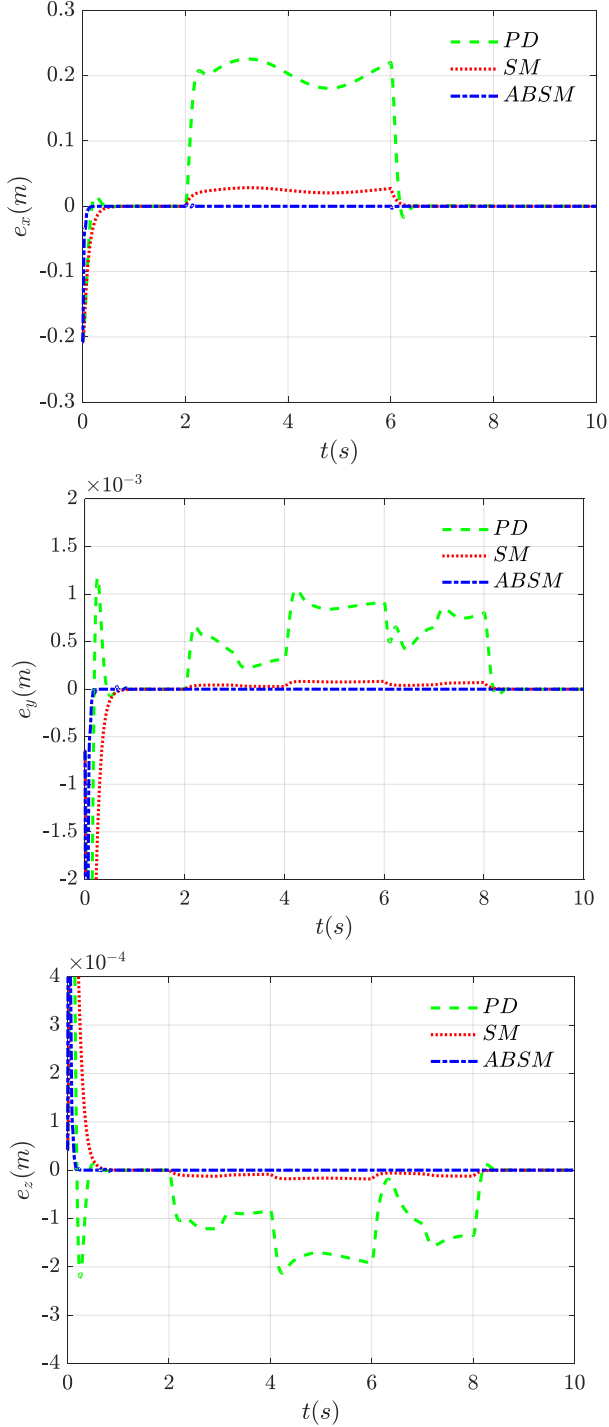

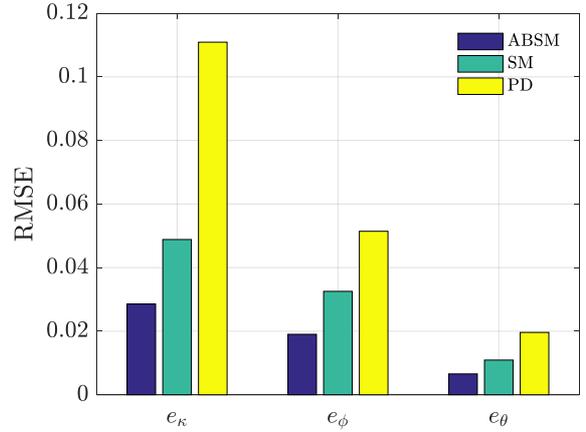

$$\text{RMSE} = \sqrt{\frac{\int_{t_0}^{t_1} e_\Omega^2 \, dt}{t_1 - t_0}} \tag{48}$$

where $e_\Omega$ denotes the tracking error in configuration space and $\begin{bmatrix} t_0 & t_1 \end{bmatrix}$ is the time interval of simulation. The RMSE value is calculated and plotted in Fig. 8 for three controllers. It is shown that for the proposed controller, the errors remain under 0.03, while PD controller shows the highest deviation from the desired values, showing more than 0.11 curvature error.

**Fig. 8 Error metric for comparing the tracking performance of ABSM, SM and PD controllers**

In addition, to demonstrate the effectiveness of the proposed adaptive variable impedance strategy, error criteria named as Integral Squared Error (ISE), Integral Absolute Error (IAE) and Integral Time-weighted Absolute Error (ITAE), for different controllers are investigated.

$$IAE = \int |e(t)| dt$$
$$ITAE = \int t \cdot |e(t)| dt \tag{48}$$
$$ISE = \int e(t)^2 dt$$

In what follows, two scenarios are planned. For the sake of brevity, these indices are determined only based on curvature error.

- **First scenario:**

In this scenario, the abovementioned performance indices are evaluated in the presence of 10 % parametric uncertainty (PU). As can be seen from table 4, the results of the ABSM-based variable impedance control are

**Fig. 7 Error components in task space for the controllers**

### 5.1-Comparative results

To quantitatively compare the performance of the proposed controller with the others, a root mean square error (RMSE) metric in configuration space is defined as following:

significantly better than the others. It should be noted that these indices are determined based on curvature error.

Table 4. Evaluation of performance indices -10% PU

| Controller | IAE | ITAE | ISE |
|---|---|---|---|
| PD | 0.49214 | 1.76307 | 0.20928 |
| SM | 0.25828 | 0.19254 | 0.13841 |
| ABSM | 0.05097 | 0.00231 | 0.05054 |

- **Second scenario:**

In this scenario, the abovementioned performance indices are evaluated under the influence of 25 % parametric uncertainty (PU). Similar to the previous evaluation, from table 5, the results of the proposed adaptive variable impedance control provide considerable accuracy compared to the others.

Table 5. Evaluation of performance indices -25% PU

| Controller | IAE | ITAE | ISE |
|---|---|---|---|
| PD | 0.45367 | 1.5532 | 0.20492 |
| SM | 0.2660 | 0.1937 | 0.12459 |
| ABSM | 0.05091 | 0.00234 | 0.05210 |

Fig.9 illustrates the length of the cables during simulation time. It can be noticed that the length of all cables remained constant after a very small variation at the beginning of the time interval. From the kinematic equation ( $\dot{\chi} = J\dot{q}$ ), it can be inferred that this small deviation prevents the large position error in task space. Fluctuation-free nature of these signals stems from the proposed adaptive variable impedance control strategy.

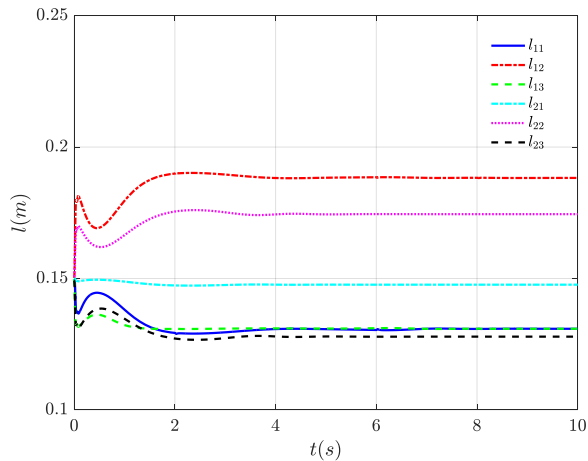

**Fig. 9 Length of the cables**

To ensure the stability of the closed-loop system, the negative definiteness property of $\dot{V}$ for the variable impedance control is depicted in Fig. 10. It is evident from this figure that during the whole simulation time period, $\dot{V}$ remains negative and holds in the vicinity of zero at the end of the simulation (Eq.32). Furthermore, it is observed that the controller guarantees the stability of the system even during the time that external loads are applied to the system.

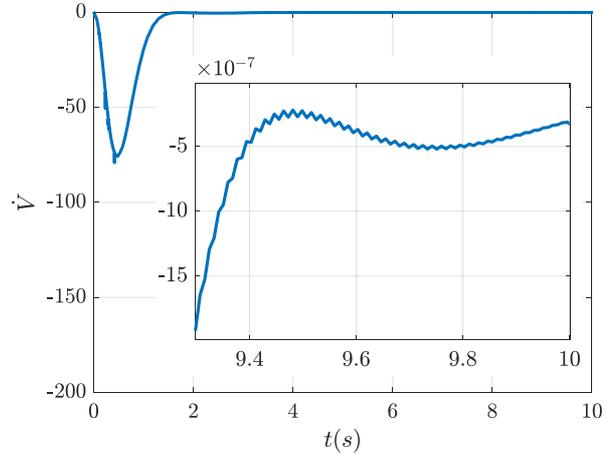

**Fig. 10 Evolution of time-derivative of Lyapunov function for variable impedance**

Phase portraits of the state variables in task space and configuration space are presented in Fig. 9(a) and 9(b). It is clearly seen that all trajectories converge to zero, which highlights the stability of the closed-loop system.

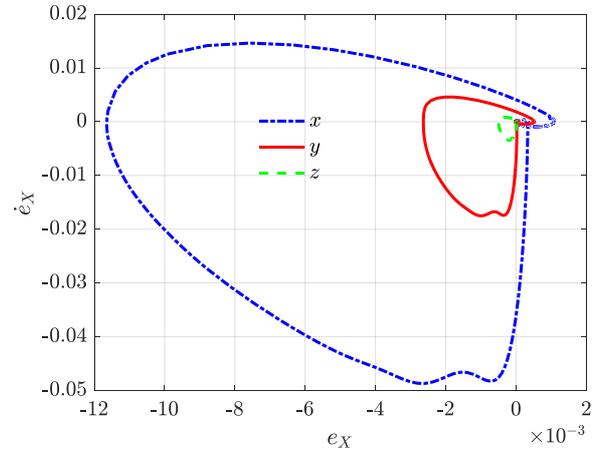

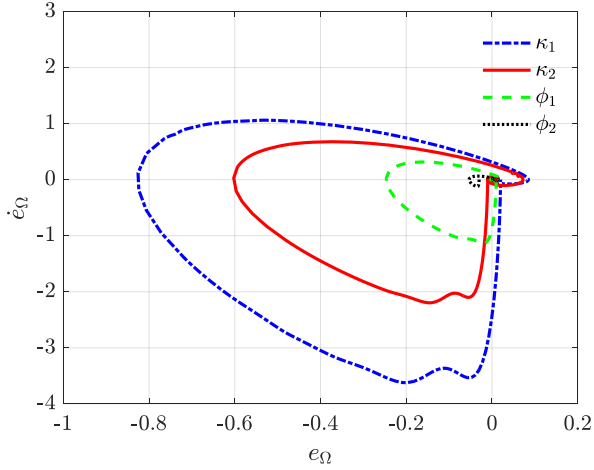

**Fig. 11 Phase portrait in (a) task space (b) configuration space**

The control commands for variable and invariable impedance control strategies are presented in Fig. 12. At the beginning of the time period, the control signals experience some fluctuations which is required to bring the robot to its desired configuration. Nonetheless, the control signals remain bounded during the entire time span. The difference between the control methods is observed in small fluctuations of variable impedance control signals, caused by desired time-varying stiffness. Another interesting point of this figure is that the signals are smooth and they are free from sharp fluctuations, which shows effectiveness of the proposed control strategy for the applications with robot-environment interaction.

The standard deviation and mean value of the control signals in configuration space for both variable and invariable impedance control methods are computed and presented in Fig 13. It can be clearly seen that by using variable impedance control, both $S_{\tau_c}$ and $MV_{\tau_c}$ are close to the corresponding values for invariable one, which means that variable impedance not only does not increase these values but also holds these values less than or close to the values of invariable impedance.

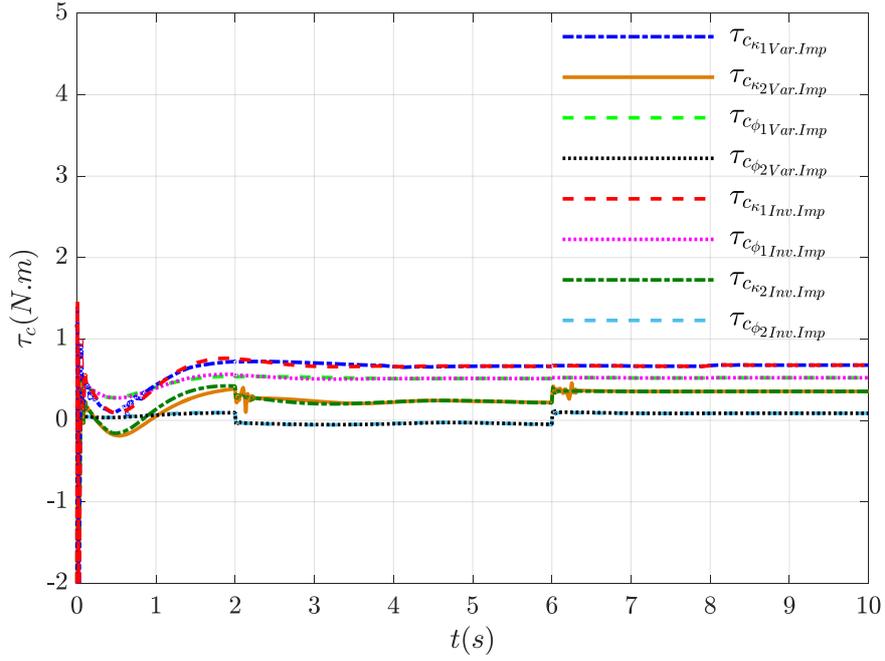

**Fig. 12 Actuation Torque for variable and invariable impedance controllers**

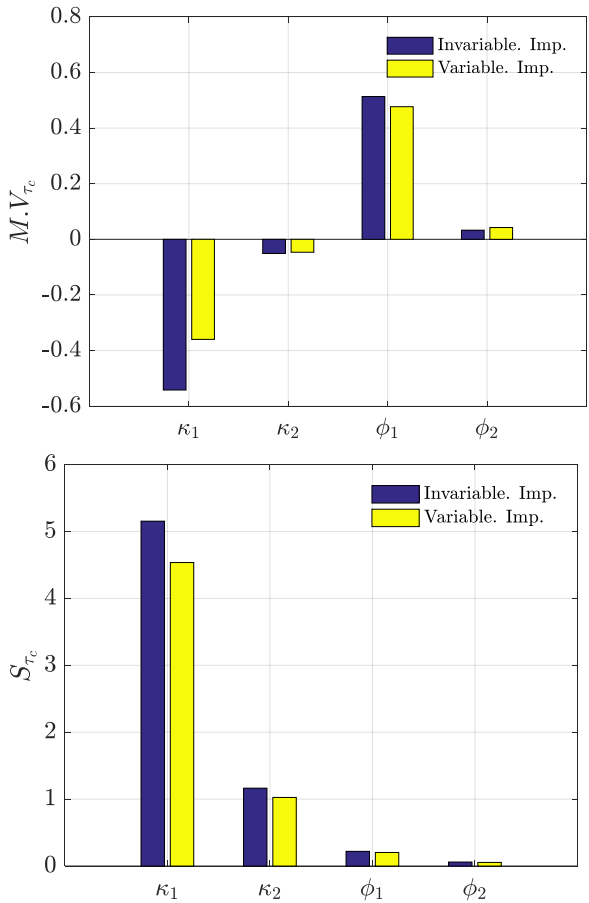

**Fig. 13 Mean value and standard deviation of actuation torques for invariable and variable impedance**

## 6- Discussion

Impedance control with constant matrices makes the closed-loop system passive. However, any arbitrary change in the impedance parameters leads to loss of the passivity. Stability is one of the most important issues in control systems, specifically when there is an interaction with the environment. Thus, in order to ensure the stability of the impedance system with time-varying parameters, in this study, we imposed some constraints to guarantee the stable behavior of the impedance system in the presence of variable inertia, damping, and stiffness profiles for the MSRM. The impedance control needs to find a compromise between the tracking accuracy and the physical interaction with the environment, which means that a perfect positioning accuracy is not the control objective. This issue was addressed in this study by the proposed variable impedance control. It is observed that the error converges to zero in both configuration and task spaces, thereby ensuring the stability of the impedance system (Fig.7 and Fig.11). It should be noted that some other profiles exist that do not satisfy the constraints although they allow qualitatively stable behavior, and that is why the constraints are conservative to guarantee the stability.

Due to the complexity of measuring the external forces and torques in real applications of soft robots, the load estimator can be applied; indeed, its accuracy depends upon the position and velocity of the configuration states. Since the relevant states in joint or actuator spaces can be
15

indirectly measured, it is evident that the kinematic error can decrease the effectiveness of the estimator. Consequently, this study aimed to propose a variable impedance control in configuration space based on the derived dynamic model of the fluidic-tendon driven soft robot in configuration space in the presence of external loads (Fig.6). For soft robots, it is possible to have multiple contacts with the environment, which impose to determine the kinematics of these points in impedance control. To tackle this issue, the momentum-based load estimator was applied in the configuration space.

In spite of time-varying impedance matrices, the control commands were free from fluctuations and variations thanks to the presented solution. As anticipated, relatively small fluctuations of errors in both task and configuration spaces lead to small variations in the length of the cables, which confirm the effectiveness of the variable impedance control (Fig.9). In addition, due to variable stiffness, it is expected to have a fluctuation pattern for actuation torques while the proposed strategy is free from this and a very small difference between actuation torques can be seen (Fig.12).

## 7- Conclusion

In this paper, a new adaptive variable impedance control strategy was designed for a modular soft-robot manipulator by integrating an adaptive back-stepping sliding mode control and a load estimator in configuration space. Uncertainties were estimated by integrating an adaptation law. The stability of the closed-loop system was addressed by using the Lyapunov theory. Effectiveness of the proposed control structure was demonstrated through simulation results. The performance of this control strategy was compared with the performance of two nonlinear controllers, an inverse dynamics PD control and a conventional sliding mode control. It was observed that the external forces could be estimated using an effective estimation law, and their effects were compensated through the varying impedance structure. Results validated the assumptions on the variable impedance control of ensuring a compromise between the force and position in configuration space. Real-time implementation and incorporation with a fault-tolerant control strategy would be interesting as a future works.

Compliance with Ethical Standards:
(in case of Funding) Funding: This study was not funded by anyone.

Conflict of Interest: The authors declare that they have no conflict of interest.